\documentclass{article}
\usepackage{spconf,amsmath}
\usepackage[final]{graphicx}
\usepackage{subcaption}
\usepackage{xcolor}
\usepackage{multirow}
\usepackage{multicol}
\usepackage[final]{hyperref} % adds hyper links inside the generated pdf file

\hypersetup{
	colorlinks=true,       % false: boxed links; true: colored links
	linkcolor=blue,        % color of internal links
	citecolor=blue,        % color of links to bibliography
	filecolor=magenta,     % color of file links
	urlcolor=blue         
}
\title{Deep learning based detection of  Acute Aortic Syndrome in contrast CT images}
\makeatletter
\def\@name{ \emph {Manikanta Srikar Yellapragada}$^{\star \dagger }$, 
\emph {Yiting Xie}$^{\dagger}$,
\emph {Benedikt Graf}$^{\dagger}$, 
\emph {David Richmond}$^{\dagger}$,
\\ \emph {Arun Krishnan}$^{\dagger}$, 
\emph {Arkadiusz Sitek}$^{\dagger}$}
\makeatother

\address{\\$^{\star}$ New York University, New York, NY, USA \\ $^{\dagger}$IBM Watson Health, Cambridge, MA, USA}

\begin{document}
\ninept

\maketitle
\begin{abstract}
Acute aortic syndrome (AAS) is a group of life threatening conditions of the aorta. We have developed an end-to-end automatic approach to detect AAS in  computed tomography (CT) images. Our approach consists of two steps. At first, we extract $N$ cross sections along the segmented aorta centerline for each CT scan. These cross sections are stacked together to form a new volume which is then  classified using two different classifiers, a 3D convolutional neural network (3D CNN) and a multiple instance learning (MIL). We trained, validated, and compared two models on 2291 contrast CT volumes. We tested on a set aside cohort of 230 normal and 50 positive CT volumes. Our models detected AAS with an Area under Receiver Operating Characteristic curve (AUC) of $0.965$ and $0.985$ using 3DCNN and MIL, respectively.
\end{abstract}
\begin{keywords}
Acute aortic syndrome, dissection, CT imaging, deep learning, radiology reports
\end{keywords}
%
% \vspace{-0.2cm}

\section{Introduction}
\label{sec:intro}

\begin{figure}[ht!]
    \centering
    \includegraphics[trim={0.2cm 0.3cm 0.3cm 0cm},clip,width=8cm]{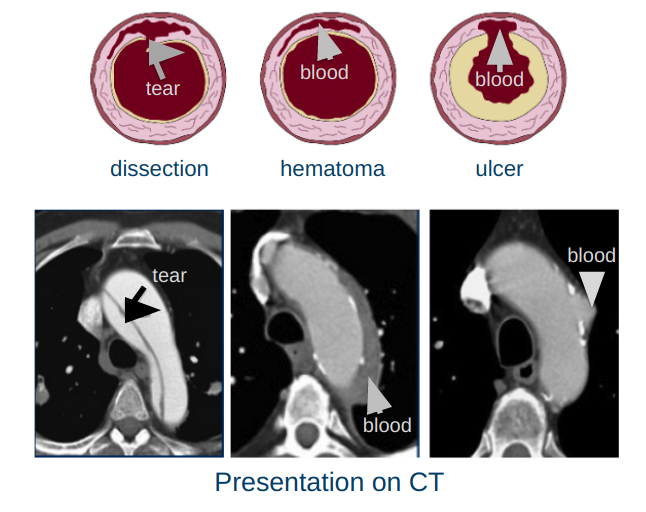}
    % \caption{Types of Acute aortic syndromes. Source \cite{baliga2014role}.}
% \label{fig:aas}
% \end{figure} 
% \vspace{-0.6cm}
% \begin{figure}[ht!]
    % \centering
    % \begin{subfigure}[t]{0.15\textwidth}
    %     \centering
    %     \includegraphics[width = 2.5cm, height = 3cm]{images/ct_d.png}
    %     \caption{Aortic dissection}
    % \end{subfigure}%
    % ~ 
    % \begin{subfigure}[t]{0.15\textwidth}
    %     \centering
    %     \includegraphics[trim={0 0.5cm 0 0.2cm},clip,width = 2.5cm, height = 3cm]{images/ct_h.png}
    %     \caption{ Hematoma }
    % \end{subfigure}
    % ~
    %         \begin{subfigure}[t]{0.15\textwidth}
    %     \centering
    %     \includegraphics[trim={0 0.5cm 0 0.2cm},clip,width = 2.5cm, height = 3cm]{images/ct_u.png}
    %     \caption{Penetrating ulcer }

    % \end{subfigure}
    \caption{Types of Acute aortic syndrome. Source \cite{baliga2014role}, \cite{RadiologyAssistant}}

\label{fig:ct_aas}
\end{figure}

Acute aortic syndromes (AAS) are a constellation of life threatening medical conditions including aortic dissection, intramural hematoma (IMH) and penetrating atherosclerotic aortic ulcer \cite{corvera2016acute}.  Acute aortic dissection is the most frequent among the three, which is defined as a separation of the layers of the aortic wall due to an intimal tear. Aortic dissection is a serious event associated with a high mortality. Untreated death rates of 40\% on initial presentation and increase $>$1\% per hour have been reported \cite{sarasin1996detecting}. Aortic intramural hematoma represents a variant of dissection characterised by the absence of an entrance tear. Penetrating atherosclerotic aortic ulcer describes the condition in which ulceration of an aortic atherosclerotic lesion penetrates the internal elastic lamina into the media1.  The difference between the three types of AAS is shown in Figure \ref{fig:ct_aas}. CT with contrast is typically used for diagnosis of AAS.

Convolutional Neural Networks (CNNs) have been established as a powerful class of methods for understanding image and video content, achieving state of the art results on image classification \cite{zoph2016neural}, segmentation \cite{chen2017deeplab}, detection \cite{he2017mask} and in medical image analysis \cite{ronneberger2015u}, \cite{sarasin1996detecting}. We developed an end-to-end approach to detect AAS on volumetric CT images using CNNs. This can be used to prioritize worklist in order to make the diagnosis and treatment be accomplished more quickly or send alerts to the radiologists or other clinicians that the study needs immediate attention or review.\\

% \vspace{-2mm}
\section{Related Work}
\label{sec:related}
Dehghan et al. \cite{dehghan2017automatic} developed a method for patient level detection of aortic dissection by segmenting the aorta to obtain cross section images, followed by flap detection and shape analysis to detect dissection in the slices. They tested their model on 37 contrast-enhanced columns achieving an accuracy of 83.7 \%. However, they didn't make use of deep learning methods in segmentation and detection and used a small set for testing. 

Harris et al. \cite{robertsiim} developed a fast method to detect aortic dissection in CT images in which they extracted the center patch from each slice, passed it through the CNN and used the fraction of total slices positive as a metric to decide if the overall volume was positive. Xu et al. \cite{xu2019automatic} followed a similar approach as ours where the aorta region was segmented by deep learning, processed and classified. However, they made use of slice level ground truth, which would take a lot of effort to curate. Our method only uses volume level labels, which can be easily obtained from radiology reports. 

\section{Proposed Approach}
\label{sec:approach}
\subsection{Dataset and preprocessing}
We used a dataset comprising 20,000 CT scans (from 20000 subjects) and associated radiology reports. Radiology reports are a written communication between the radiologist interpreting the CT image and the physician who requested the examination. We extracted the following keywords from the Findings and Impression sections of the radiology reports - ``dissection'', ``hematoma,'' and ``ulcer''. We found that 838 reports had mentions of those keywords. We then manually examined the reports and annotated them as positive, uncertain, or negative. This provided us with 500 positive (class 1) cases. The remaining 338 scans were discarded as most of them were uncertain due to motion artifacts and poor visualization. To create normal cohort (class 0), we randomly selected 3000 contrast CT scans/reports from the reports with no mentions of keywords related to AAS. 

\subsection{Aortic centerline segmentation}
\begin{figure}[ht!]
    \centering
    \includegraphics[ width=8cm]{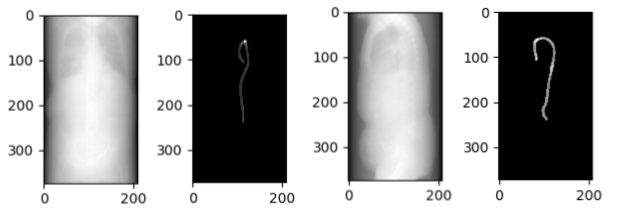}
    \caption{Centerline of aorta. Left - Coronal, Right - Sagittal}
\label{fig:aorta}
\end{figure}

We applied a previously developed deep learning based aortic centerline segmentation algorithm \cite{AneurysmRSNA} on all CT scans in training, validation, and test sets. This algorithm segmented the approximate centerline inside of the aortic lumen. We performed the localization and straightening of aorta as a first step because we wanted the algorithm to focus on the relevant anatomy (i.e. the aorta) and ignore other regions. Furthermore, we wanted it to be less prone to bias caused by other artifacts present in the images. Figure \ref{fig:aorta} shows the centerline of the aorta in coronal and saggital view. The segmented centerline served two purposes. First, it was used to construct the image subvolume used for AAS detection; Second, it was used to find average Hounsfield Unit (HU) in the center of the aorta which was used match distributions of normal and abnormal scans in all datasets (see section~\ref{label:matching}).

% \vspace{-0.5cm}
\subsection{CT contrast/intensity matching}
\label{label:matching}
\begin{figure}[ht!]
    \centering
    \begin{subfigure}[t]{0.24\textwidth}
        \centering
        \includegraphics[width=4cm]{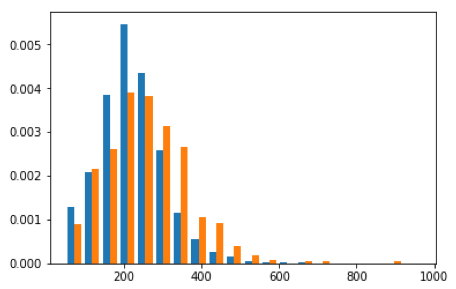}
        \caption{3000 Negatives, 500 Positives }
    \end{subfigure}%
    ~ 
    \begin{subfigure}[t]{0.24\textwidth}
        \centering
        \includegraphics[width=4cm]{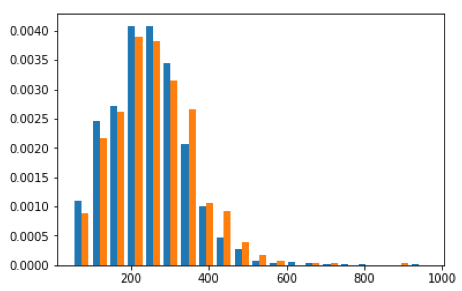}
        \caption{2100 Negatives, 500 Positives}
    \end{subfigure}
    \caption{Histogram of mean HU before and after matching by removal.}
    \begin{tabular}{r@{: }l r@{: }l}
X-axis & mean HU & Y axis & ratio of scans\\
Orange& Positive & Blue& Negative 
\end{tabular}
\label{fig:hist}
\end{figure}

Using the segmented centerline, we constructed the image volumes and calculated the mean HU value of the aorta region in each volume. To ensure that the distribution of aorta mean HU value was the same between positive and negative classes, we discarded 900 negative cases using a histogram-bin matching method. We only selected a subset of the normal cases so that the algorithm would not pick up the difference in contrast/intensity between the normal and dissection cases. We used a greedy algorithm  where at each iteration we removed negative scans from the histogram bin with the maximum discrepancy between positive and negative histograms. The iterations were terminated when the cross correlation exceeded a predefined threshold. This approximate equalization of histograms was done to ensure that the deep learning methods would not utilize the difference in HU value between the positive and negative scans during the classification. The histograms before and after application of the algorithms are shown in figure \ref{fig:hist}. 

% \vspace{-0.7cm}

\subsection{Extraction of cross-sectional images}
We extracted $N$ cross sectional images perpendicular to and along the segmented centerline of the aorta using a previously developed method. The extraction was performed in an isotropic image space, which had a high xyz resolutions of typical pixel size of 0.7mm$^3$ to capture the fine details of the aorta. Each cross sectional patch contained the aorta as well as some background information. We chose patch size  to be 114x114 pixels because with 0.7 mm it covered the largest diameter aortas in our dataset. The extracted patches from each CT scan were stacked together to form a `straightened aorta'. The number of patches varied between 150 and 1000 due to the varying length of the aorta in the CT scan.  The image intensity of each extracted straightened aorta subvolume was clipped between -1024HU and 2048HU, and then scaled down to $[0,1]$. Fifty slices were extracted randomly from the volume and each slice's neighbouring slices were used, i.e. the dimension of the input to the CNN was (50,114,114,3). 

We decide to use 50 cross-sections slices to have sequence long enough to cover the entire aorta and at the same time short enough for LSTM temporal memory model to be effective. We have not optimized this number which is future work. Although only 50 slices are selected, this section is random for every epoch
(and augmentation) and therefore in principle all slices contribute information during training.

In total, 450 out of 500 positive and 1841 out of 2100 normal volumes were used for training, with a train - validation split of 80-20. Our test data consisted of 50 positive and 230 negative CT scans not used in training. Some patients had more than one scan and it was made sure that there was no data leak between data sets {i.e.} volumes of the same patients were always in the same set. 

\subsection{Machine Learning Models for AAS Classification}
Two different deep learning (DL) methods were used for the detection of AAS. They were I3D + ConvLSTM and Multiple Instance Learning (MIL) model. The justification for the first architecture is that short term spatio temporal features would be learned by the 3D CNN and long term spatiotemporal features are captured by bidirectional ConvLSTM. We use the MIL based approach as the slice level sigmoid outputs would correspond to the probability of that slice having the disease, which are then combined by the adaptive pooling method to obtain the final output. These methods also doesn't require local annotation of AAS manifestations. The first approach, I3D + ConvLSTM is a 3D approach as it uses sequences of 2D images (2D + time), whereas the MIL approach is 2D. 
\subsubsection{I3D + ConvLSTM}
\begin{figure}[ht!]
    \centering
    \includegraphics[width=8cm]{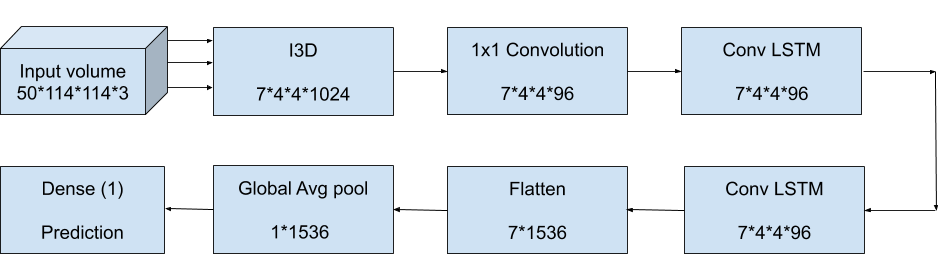}
    \caption{Overview of I3D + ConvLSTM architecture.}
\label{fig:i3d}
\end{figure}

The architecture was composed of inflated inception model, followed by two Convolutional Long Short Term Memory (LSTM) layers, average pooling and dense layers as shown in figure \ref{fig:i3d}. The architecture was inspired from \cite{zhang2017learning} in which 3DCNN followed by ConvLSTM was used for Gesture recognition. ConvLSTM is a variant of LSTM containing the convolution operation inside the LSTM cell. ConvLSTM replaces matrix multiplication with convolution operation at each gate in the LSTM cell. By doing so, it captures underlying spatial features by convolution operations in multiple-dimensional data. It was first introduced in \cite{xingjian2015convolutional}.

We used the I3D model pretrained on Imagenet and Kinetics datasets. Pretrained model was used because the model had more than 13 million parameters and training from scratch would require large amounts of data in order to avoid model over-fitting. Features from the I3D model were extracted after the last inception block and the dimensions were of $7*4*4*1024$. Seven was the temporal dimension, which was the number of slices in our case, $4*4$ was the spatial dimension, and 1024 was the number of channels.  These were passed through a 1x1 convolution layer which reduced the number of channels from 1024 to 96. This was done to reduce the number of parameters. Such extracted features were then passed through two bidirectional ConvLSTM layers.

% \vspace{-0.5cm}

\subsubsection{Multiple Instance Learning}
\begin{figure}[ht!]
    \centering
    \includegraphics[width=8cm]{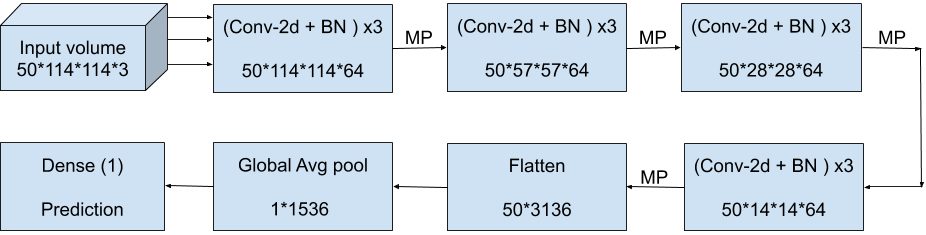}
    \caption{Overview of Multiple instance learning architecture.}
\label{fig:mil}
\end{figure}

We implemented a multiple instance learning (MIL) based model, in which each slice was passed through a 2D CNN to obtain a local label, which were then pooled to obtain the label for the entire volume. The architecture was inspired from \cite{braman2018disease} in which MIL was used to detect emphysema. 2D CNN is shown in figure \ref{fig:mil}. It consisted of four blocks, each block having three convolutional layers with 64 filters of size 3x3, followed by Batch Norm layers. Output of each block was passed through a max pool layer of size 2x2. To combine the slice level volumes, we used a trainable adaptive pooling method introduced in \cite{mcfee2018adaptive}, and was defined as 
\begin{equation}
    P_\alpha (Y | X) = \sum\limits_{x \in X} p(Y|x) \Bigg( \frac{\exp (\alpha  \cdot  p(Y|x))}{\sum\limits_{z \in X} \exp (\alpha \cdot p(Y|z))}  \Bigg)
    \label{eq:adaptive}
\end{equation}

Here $p(Y|x)$ denotes the prediction of each slice $x$, and $X$ denotes the image volume. $P_a (Y | X)$ was the volume level prediction and $\alpha$ was a trainable parameter. When $\alpha = 0$, this represents the mean of slice predictions, when $\alpha = 1$, equation \ref{eq:adaptive} is reduced to softmax pooling and when $\alpha \to \infty$, it approximates the max pooling operation. During our experiments, we found that the value of $\alpha$ after training was $1.4$ and therefore the optimal adaptive pooling was somewhere between softmax and max operations. 

All models were trained until performance on the validation set was optimal. We used a batch size of 16 for I3D and 8 for MIL. We used the Adam optimizer with a learning rate of 0.0003 for all the experiments. The models were trained on one NVIDIA Tesla V100 GPU.
% \vspace{-0.6cm}
\section{Experiments and Evaluation}

Input to the machine learning models was a 3D volume of dimension $114\times114\times N$, where  $N$ was the number of slices in axial view of the CT volume after extracting the cross sections along the aorta, $N \in [150,1000]$ . Let $X$ axis be width of the slice (114), $Y$ axis be height of the slice (114) and $Z$ axis be the number of slices ($N$). When the volume from $YZ$ and $XZ$ direction was given as the input to the 3d CNN, we found an improvement in performance. 

We trained the I3D + ConvLSTM model with data taken from XY, YZ and XZ directions and MIL model with data taken only from XY direction. All models were trained on 360 positive and 1473 negative scans and validated on 90 positive and 368 negative scans. Based on validation loss, we selected the best performing model to evaluate on the test data, which consists of 50 positive and 230 negative scans. We used binary cross entropy loss during training and AUC to evaluate the models on test data.  
% \vspace{-0.4cm}

\section{Results}
\begin{figure}[ht!]
    \centering
    \begin{subfigure}[t]{0.25\textwidth}
        \centering
        \includegraphics[trim={0 0 0 0.75cm},clip,width = 4cm]{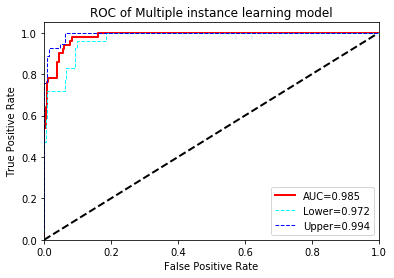}
        \caption{MIL, XY direction}
    \end{subfigure}%
    ~ 
    \begin{subfigure}[t]{0.25\textwidth}
        \centering
        \includegraphics[trim={0 0 0 0.75cm},clip,width = 4cm]{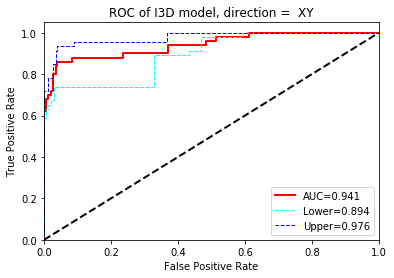}
        \caption{I3D, XY direction }
    \end{subfigure}
    
        \begin{subfigure}[t]{0.25\textwidth}
        \centering
        \includegraphics[trim={0 0 0 0.75cm},clip,width = 4cm]{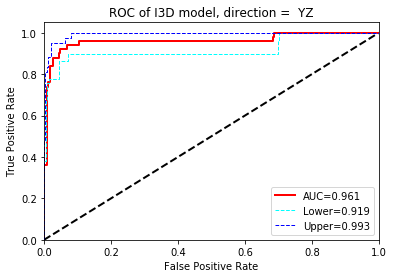}
        \caption{MIL, YZ direction }
    \end{subfigure}%
    ~
            \begin{subfigure}[t]{0.25\textwidth}
        \centering
        \includegraphics[trim={0 0 0 0.75cm},clip,width = 4cm]{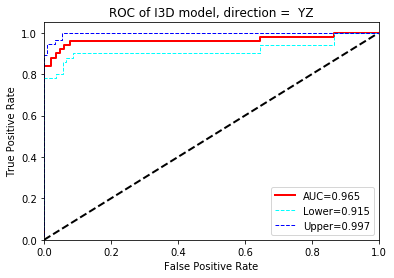}
        \caption{I3D, YZ direction }
    \end{subfigure}
    
            \begin{subfigure}[t]{0.25\textwidth}
        \centering
        \includegraphics[trim={0 0 0 0.75cm},clip,width = 4cm]{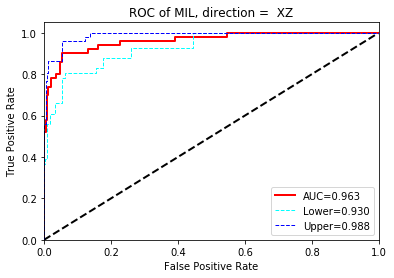}
        \caption{MIL, XZ direction }
    \end{subfigure}%
    ~
            \begin{subfigure}[t]{0.25\textwidth}
        \centering
        \includegraphics[trim={0 0 0 0.75cm},clip,width = 4cm]{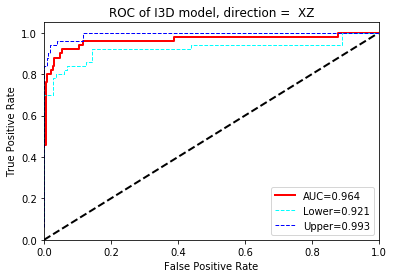}
        \caption{I3D, XZ direction }
    \end{subfigure}
    \caption{ROC curves with 95 \% confidence interval.}

\label{fig:roc}
\end{figure}

I3D + ConvLSTM based model demonstrated achieved an AUC = 0.945. Multiple instance learning (MIL) + adaptive pooling based method outperformed the I3D model, achieving an AUC = 0.985. ROC curves with 95 \% confidence interval of MIL and I3D models with XY, YZ and XZ direction data are shown in figure \ref{fig:roc}.

% \begin{figure}[ht!]
%     \centering
%     \begin{subfigure}[t]{0.25\textwidth}
%         \centering
%         \includegraphics[trim={0 0 0 0.75cm},clip,width=4cm]{images/mil_roc.png}
%         \caption{MIL, XY direction}
%     \end{subfigure}%
%     ~ 
%             \begin{subfigure}[t]{0.25\textwidth}
%         \centering
%         \includegraphics[trim={0 0 0 0.75cm},clip,width=4cm]{images/i3d_yz_roc.png}
%         \caption{I3D, YZ direction }
%     \end{subfigure}
    
%     \caption{ROC curves of best performing models with 95 \% confidence interval.}

% \label{fig:roc}
% \end{figure}

\begin{table}[ht]
\centering
\begin{tabular}{|c|c|c|c|c|}
\hline
\multirow{2}{*}{Model}     & \multirow{2}{*}{\# Parameters} & \multicolumn{3}{c|}{AUC}          \\ \cline{3-5} 
                          &                                & Mean  & Lower CI & Upper CI \\ \hline
I3D  - XY        & 13,391,345                     & 0.941 & 0.894       & 0.976       \\ \hline
I3D - YZ        & 13,391,345                     & 0.964 & 0.921       & 0.993       \\ \hline
I3D - XZ        & 13,391,345                     & 0.965 & 0.915       & 0.997       \\ \hline
MIL - XY & 414,210                        & 0.985 & 0.972       & 0.994       \\ \hline
MIL - YZ & 414,210                        & 0.961 & 0.919       & 0.993       \\ \hline
MIL - XZ & 414,210                        & 0.963 & 0.930       & 0.988       \\ \hline

\end{tabular}
\caption{AUCs with 95\% confidence interval.}
\label{table:results}
\end{table}

\section{Discussion and Conclusion}
We have developed an end-to-end deep learning based approach to detect Acute Aortic Syndrome in volumetric contrast CT images. We obtained performance in detecting presence of AAS reaching 0.985 AUC. We tested I3D+ConvLSTM and Multiple Instance  Learning approaches and they both performed comparably with MIL performing only slightly better. We find an improvement in performance in the I3D model when we reslice the data in three orthogonal directions (x,y,z) and train three classifiers separately for each direction. We follow by averaging resulting predictions from those three classifiers to obtain overall score. We speculated that those different directions contain slightly different spatial information and predictions will not be fully correlated. There was no significant improvement when this reslicing technique was used for MIL algorithm. 

The study has limitations. We cannot guarantee that positive and negative data distributions are identical.  Although we specifically addressed this issue by equalizing distribution of HU values in aorta (section~\ref{label:matching}) there may be some differences remaining. We also completely rely on radiology reports to provide the ground truth and therefore rely on single radiologist opinion when generating the ground truth.

\bibliographystyle{IEEEbib}
\bibliography{Template_ISBI}

\end{document}